\documentclass[10pt,twocolumn]{article}
\usepackage[margin=0.72in,columnsep=0.23in]{geometry}
\usepackage{times}
\usepackage{microtype}
\usepackage{graphicx}
\usepackage{booktabs}
\usepackage{multirow}
\usepackage{multicol}
\usepackage{amsmath,amssymb,bm}
\usepackage{algorithm}
\usepackage[noend]{algpseudocode}
\usepackage{xcolor}
\usepackage{url}
\usepackage[hidelinks]{hyperref}
\usepackage[font=small,labelfont=bf]{caption}
\usepackage{subcaption}
\usepackage{enumitem}
\usepackage{float}
\usepackage{placeins}
\usepackage{xspace}


\definecolor{grpo}{HTML}{0077BB}
\definecolor{fire}{HTML}{CC3311}
\definecolor{sft}{HTML}{EE7733}
\definecolor{gt}{HTML}{009988}
\newcommand{\method}{FIRE-VLA\xspace}

\title{FIRE-VLA: Failure-Informed Self-Evolution for\\
Vision-Language-Action Models in Autonomous Driving}
\author{Hao Dou\\
Harbin Institute of Technology}
\date{}

\hypersetup{
    pdftitle={FIRE-VLA: Failure-Informed Self-Evolution for Vision-Language-Action Models in Autonomous Driving},
    pdfauthor={Hao Dou}
}

\begin{document}
\maketitle

\begin{abstract}
Reinforcement learning improves autonomous-driving vision-language-action (VLA) models by evaluating trajectories sampled from the current policy. Group relative policy optimization (GRPO) learns from reward differences within each rollout group. When all sampled trajectories are poor, this relative signal can rank failures without identifying behavior outside the failed region. We introduce \method, a failure-informed self-evolution framework that converts such unresolved failures into privileged supervision for the next policy. Low-reward, low-diversity groups trigger self-distillation from a frozen round-start copy of the same model. Teacher and student have the same parameter scale, but only the teacher observes the hidden future trajectory. Supervision follows the student's generated prefix and is restricted to answer tokens, while GRPO remains active for every group. The updated policy supplies the teacher for the next round, allowing the routed failure distribution to change with the policy without requiring a larger external teacher. Starting from the same Qwen2.5-VL-3B SFT checkpoint, the comparison matches student rollout and policy-update counts. On 6,019 examples from 150 held-out nuScenes scenes, \method retains comparable single-sample planning, reduces $G{=}4$ mean L2 from 1.848 to 1.500~m, and lowers evaluation-persistent failure prevalence from 13.03\% to 11.20\%. The reduction in mean error arises mainly from rare severe rollouts rather than uniform improvement across ordinary trajectories.
\end{abstract}


\begin{figure*}[t]
    \centering
    \includegraphics[width=.98\textwidth]{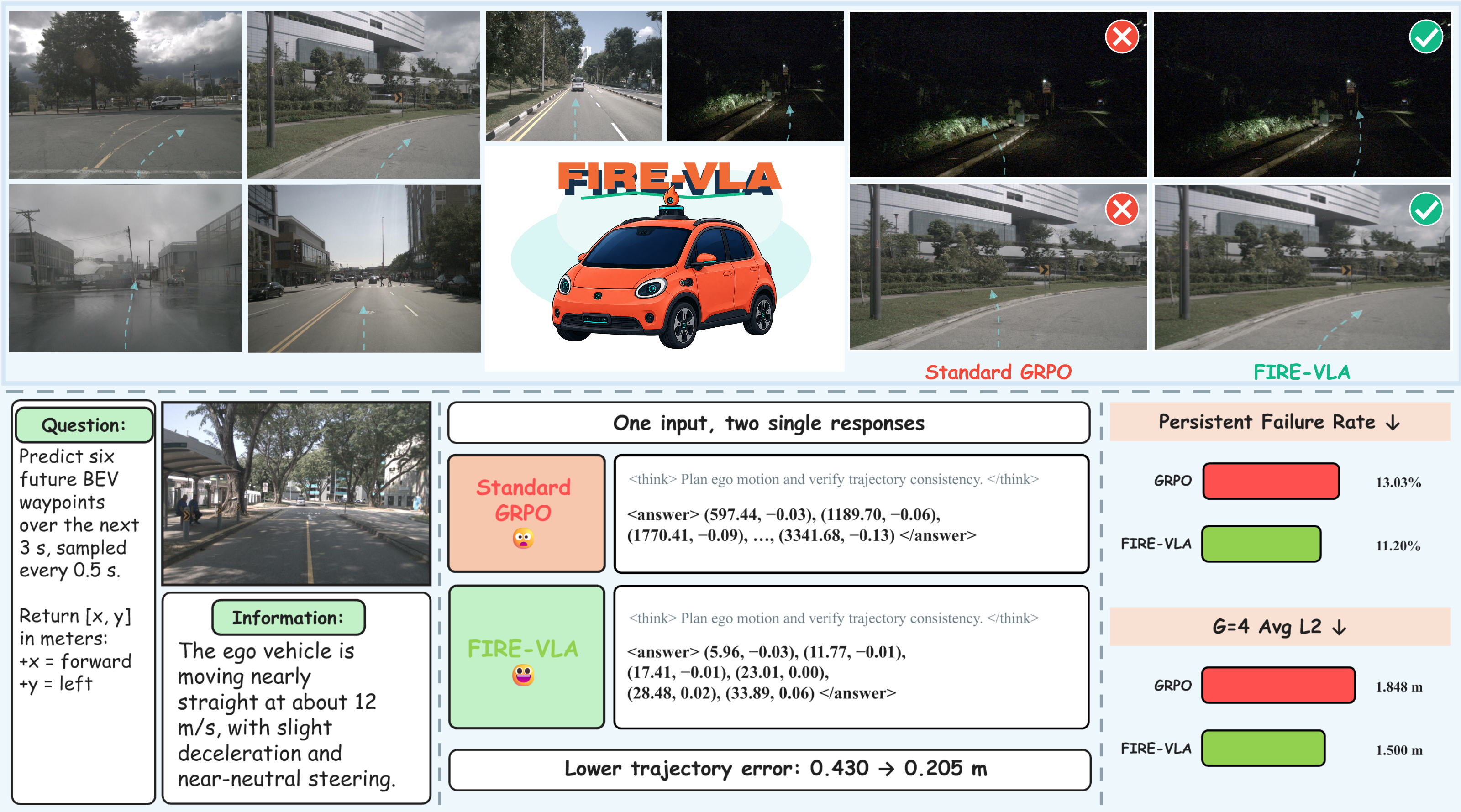}
    \caption{\textbf{FIRE-VLA at a glance.} Top: representative nuScenes scenes and paired trajectory contrasts between standard GRPO and FIRE-VLA. Bottom: a front-view image and historical ego motion condition six future BEV waypoints. The paired stochastic response shows an implausibly scaled GRPO trajectory and a plausible FIRE-VLA trajectory; the separate callout gives a normal-scale single-sample comparison. Across all 6,019 fixed samples, FIRE-VLA reduces evaluation-persistent failures from 13.03\% to 11.20\% and $G{=}4$ Avg.~L2 from 1.848 to 1.500~m. The cases are illustrative; Sec.~4 and Appendix~\ref{app:qualitative} specify the fixed selection protocol and paired analysis.}
    \label{fig:overview}
\end{figure*}

\section{Introduction}

Vision-language-action (VLA) models connect visual perception and language reasoning to executable driving behavior. Recent systems use this interface for instruction following, graph-structured scene understanding, closed-loop control, and numerical trajectory generation \cite{shao2024lmdrive,sima2024drivelm,pan2024vlp,fu2025orion}. Supervised fine-tuning (SFT) supplies driving priors and a structured action format. Reinforcement learning (RL) can then improve the policy from trajectories sampled from and evaluated under its own on-policy response distribution. AutoDrive-R$^2$, for example, demonstrates the value of GRPO-style post-training for a reasoning-based driving VLA \cite{yuan2026autodriver2}.

Group relative policy optimization (GRPO) derives its learning signal from reward contrast among multiple responses to one prompt \cite{shao2024deepseekmath}. A group containing both successful and poor trajectories provides an informative ranking of sampled behavior. Not every failure group has this structure. In an \emph{unresolved failure group}, all trajectories remain poor and produce similar reward outcomes. GRPO can rank these failures without identifying how to escape the failed region. The limitation is insufficient corrective information, not the disappearance of the GRPO objective or a claim that its gradient must be small.

The missing ingredient is corrective information that is specific to the policy's own failed region. Existing approaches obtain such information through failure refinement or privileged distillation \cite{luo2026elfvla,zhao2026opsd}. We study whether the same VLA can instead use its unresolved failures to supervise its successor, without introducing a larger external teacher. Figure~\ref{fig:overview} presents the trajectory-prediction interface, representative responses, and the aggregate outcomes used in our evaluation. The examples are illustrative; the quantitative claims come from the paired 6,019-sample evaluation.

We introduce \textbf{FIRE-VLA}, a failure-informed self-evolution framework for VLA post-training. The method routes low-reward, low-diversity rollout groups to privileged self-distillation while retaining GRPO for all groups. A frozen copy of the round-start policy serves as the teacher. It has the same parameter scale as the student and differs only in access to the hidden future trajectory. Teacher predictions are evaluated along the student's on-policy prefix and supervise only trajectory-answer tokens. At the round boundary, the updated policy initializes both branches of the next round. Policy updates consequently change the unresolved-failure distribution that determines where privileged supervision is applied.

We compare FIRE-VLA with standard GRPO from the same Qwen2.5-VL-3B SFT checkpoint. Both methods use the same 1,200 unique RL prompts, 4,800 student rollouts, and 150 policy updates. On 6,019 examples from 150 held-out nuScenes scenes, their single-sample planning performance is comparable. With four stochastic rollouts per example, FIRE-VLA reduces mean L2 from 1.848 to 1.500~m and evaluation-persistent failure prevalence from 13.03\% to 11.20\%. Distributional analysis attributes the lower mean mainly to fewer rare, severe rollout failures; FIRE-VLA does not improve the ordinary part of the error distribution uniformly.

Our contributions are:
\begin{itemize}[leftmargin=*,nosep]
    \item We formulate failure-informed post-training around unresolved rollout groups, where relative reward ranks poor trajectories but offers limited guidance for leaving the failed region.
    \item We introduce privileged on-policy self-distillation from a frozen, same-scale teacher. Only the teacher receives the hidden future trajectory, and supervision is applied along student-generated answer prefixes.
    \item Under matched student rollout and update counts, FIRE-VLA preserves comparable nominal planning while reducing evaluation-persistent failures and severe stochastic trajectory errors relative to standard GRPO.
\end{itemize}

Code, exact configurations, and the evaluation protocol are available at \url{https://github.com/forever-free1/FIRE-VLA}.


\section{Related Work}

\paragraph{Driving vision-language-action models.}
Language-conditioned driving has progressed from instruction-aware closed-loop control to unified perception, reasoning, and planning. LMDrive maps language instructions to end-to-end driving, while DriveLM organizes scene reasoning as graph visual question answering \cite{shao2024lmdrive,sima2024drivelm}. VLP uses language-mediated representations for nuScenes planning; CoVLA pairs visual, linguistic, and trajectory supervision at scale \cite{pan2024vlp,arai2025covla}. SimLingo and ORION further align semantic reasoning or language understanding with driving action generation \cite{renz2025simlingo,fu2025orion}. FIRE-VLA uses Qwen2.5-VL-3B-Instruct as its common backbone \cite{bai2025qwen25vl} and studies post-training rather than a new VLA architecture.

\paragraph{Relative-reward post-training.}
GRPO normalizes rewards across completions of the same prompt, avoiding the separate value model used by PPO-style training \cite{shao2024deepseekmath}. AutoDrive-R$^2$ couples reasoning supervision with GRPO for driving trajectory prediction, showing that evaluated on-policy rollouts can improve a driving VLA \cite{yuan2026autodriver2}. FIRE-VLA preserves that objective for all groups. Its starting point is the information content of the group: reward contrast is sufficient for diverse outcomes, whereas uniformly poor groups need an additional corrective signal.

\paragraph{Failure-aware learning and privileged self-teaching.}
ELF-VLA identifies recurring driving failures and constructs feedback-guided corrections that are verified before reinjection into RL \cite{luo2026elfvla}. On-policy self-distillation (OPSD) lets one language model serve as both student and privileged teacher, matching distributions on student-generated prefixes to reduce off-policy mismatch \cite{zhao2026opsd}. FIRE-VLA combines unresolved-failure routing, same-policy privileged supervision on student prefixes, and round-wise teacher promotion.


\section{FIRE-VLA}
\label{sec:method}

\begin{figure*}[t]
    \centering
    \includegraphics[width=.98\textwidth]{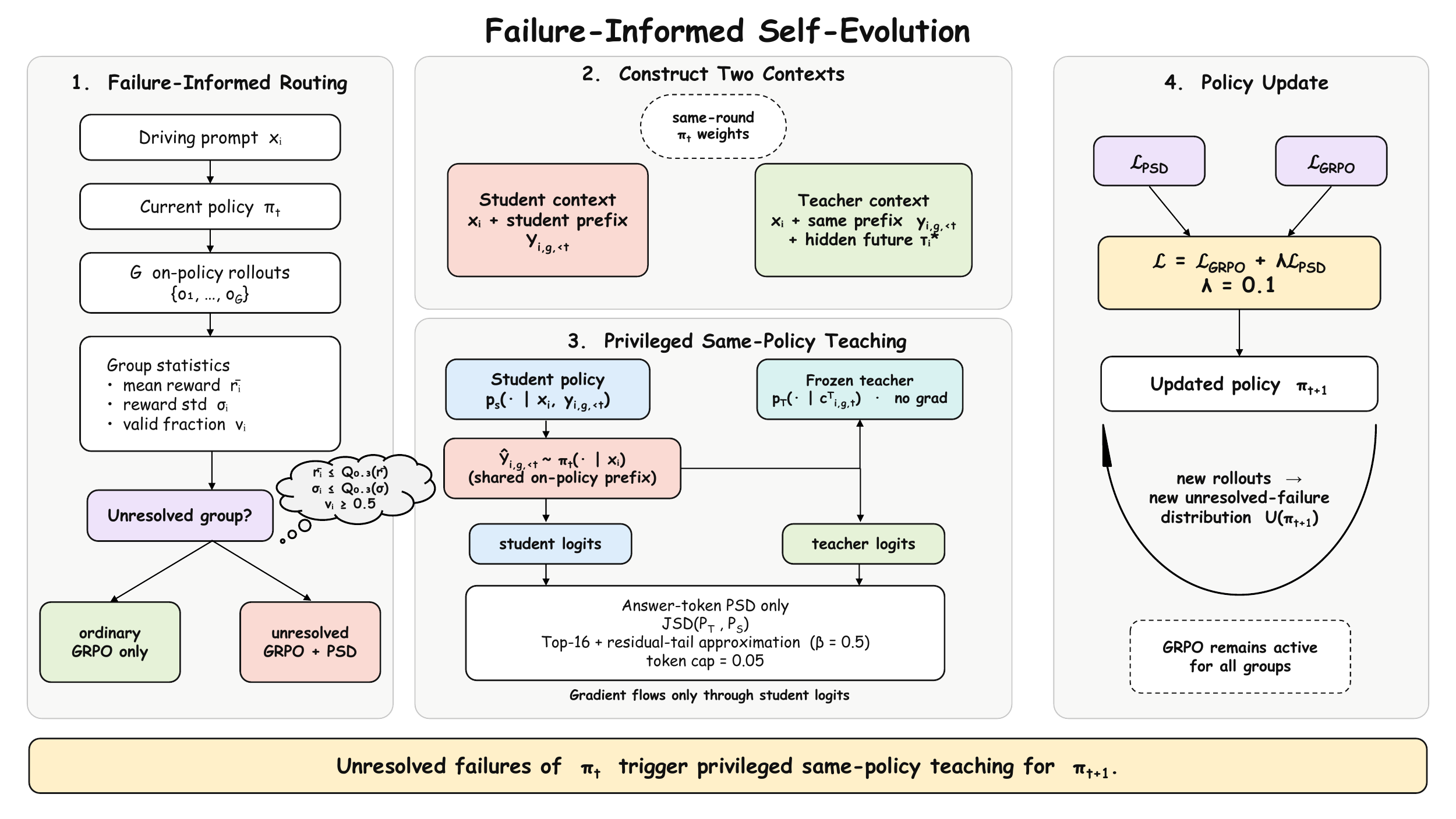}
    \caption{\textbf{Failure-informed self-evolution in FIRE-VLA.} (1) The current policy generates grouped on-policy rollouts, whose reward mean, reward standard deviation, and valid fraction route unresolved failure groups to additional PSD; GRPO remains active for every group. (2) Student and teacher start each round from the same policy and share the student's on-policy prefix, but only the teacher receives the hidden future trajectory. (3) The frozen teacher supervises the student's answer-token distribution through compact, capped JSD, with gradients flowing only through the student. (4) The routed PSD loss is combined with the universal GRPO loss to update the policy. The resulting policy induces a new rollout distribution and, consequently, a new distribution of unresolved failures for the next round.}
    \label{fig:framework}
\end{figure*}

Figure~\ref{fig:framework} summarizes one FIRE-VLA round. The method routes unresolved groups, constructs teacher and student contexts, distills answer-token distributions, and updates the policy. Both branches start from the same model; the teacher's only advantage is access to privileged future information. The updated policy also changes the failure distribution used for routing in the next round.

\subsection{Trajectory Learning with Relative Reward}

Each example contains a front-view image $I_i$, an ego-history description $h_i$, and a hidden future trajectory $\tau_i^*=(p_{i,1}^*,\ldots,p_{i,6}^*)$ at 0.5-s intervals. Given deployment input $x_i=(I_i,h_i)$, the policy produces reasoning followed by six ego-frame waypoints inside an \texttt{<answer>} span. For a valid rollout $g$, the trajectory reward used throughout formal training is
\begin{equation}
r_i^{(g)}=\left(1+\frac{1}{6}\sum_{t=1}^{6}\|p_{i,t}^{(g)}-p_{i,t}^*\|_2^2\right)^{-1},
\label{eq:reward}
\end{equation}
and invalid responses receive zero reward.

GRPO samples $G$ responses from the old policy and standardizes their rewards within each prompt group,
\begin{equation}
\hat A_i^{(g)}=\frac{r_i^{(g)}-\bar r_i}{\sqrt{G^{-1}\sum_j(r_i^{(j)}-\bar r_i)^2}+\epsilon}.
\label{eq:advantage}
\end{equation}
The clipped policy objective then reinforces responses according to these group-relative advantages \cite{shao2024deepseekmath}. When reward varies across candidates, Eq.~\ref{eq:advantage} provides useful contrast among sampled behaviors. When every response has similarly low reward, GRPO can still rank candidates within the failed region, but that ranking need not specify a behavior that exits it. FIRE-VLA therefore retains GRPO as the base objective for every group and selectively adds privileged corrective information through the four-stage process in Fig.~\ref{fig:framework}.

\subsection{Failure-Informed Routing}

Stage~1 in Fig.~\ref{fig:framework} determines where additional supervision is needed. For every on-policy group, FIRE-VLA computes reward mean $\bar r_i$, population standard deviation $\sigma_i$, and valid fraction $v_i$:
\begin{equation}
\bar r_i=G^{-1}\sum_g r_i^{(g)},\qquad
\sigma_i=\sqrt{G^{-1}\sum_g(r_i^{(g)}-\bar r_i)^2}.
\end{equation}
The \emph{training-time} route is online and batch-relative,
\begin{equation}
z_i^{\mathrm{train}}=\mathbf{1}\!\left[
\begin{array}{c}
\bar r_i\leq Q_{0.3}^{\mathrm{batch}}(\bar r)\ \land\\[-2pt]
\sigma_i\leq Q_{0.3}^{\mathrm{batch}}(\sigma)\ \land\ v_i\geq0.5
\end{array}\right].
\label{eq:train_gate}
\end{equation}
Low mean identifies poor performance, while low reward diversity marks groups with limited corrective contrast. We call a group satisfying Eq.~\ref{eq:train_gate} an \emph{unresolved failure group}. Sufficient validity prevents malformed outputs from dominating the auxiliary route. Unrouted groups receive GRPO alone; routed groups receive GRPO plus privileged self-distillation (PSD). GRPO remains active for all groups.

This online gate differs from the \emph{evaluation-time} persistent-failure detector. The latter was calibrated once on a separate 256-sample, 37-scene SFT $G{=}4$ formal validation split, with no sample or scene overlap with the final 6,019-sample evaluation. We use the same 30th-percentile convention as the training route, but freeze its numerical values on a separate SFT calibration split for cross-model evaluation. The 30th percentiles of group reward mean and population standard deviation among groups with validity at least 0.5 produced fixed thresholds $\bar r_i\leq0.4457839238$, $\sigma_i\leq0.0747977498$, and $v_i\geq0.5$. The frozen detector measures every checkpoint on a common scale, never routes training, and is never recalibrated for GRPO or FIRE-VLA.

\subsection{Privileged Future Information}

Stages~2 and 3 construct the same-policy teacher--student pair and apply privileged teaching. At evolution round $k$, student $\pi_{\theta_k}$ and teacher $\pi_{\bar\theta_k}$ begin with identical parameters. The teacher remains frozen. For rollout $g$ and its student-generated prefix $y_{i,g,<t}$, their contexts are
\begin{equation}
\begin{aligned}
c^S_{i,g,t}&=(I_i,h_i,y_{i,g,<t}),\\
c^T_{i,g,t}&=(I_i,h_i,\langle\texttt{priv}\rangle\tau_i^*\langle/\texttt{priv}\rangle,y_{i,g,<t}).
\end{aligned}
\label{eq:contexts}
\end{equation}
Teacher and student have the same parameter count and begin each round from identical weights. Only the frozen teacher sees the hidden ground-truth trajectory; the two branches otherwise share the multimodal observation and student prefix. Teacher evaluation uses no gradients, and student inputs are checked for privileged tokens. Causal masking also prevents access to future student-generated tokens. Deployment retains only the student branch.

\subsection{On-Policy Answer-Token Distillation}

As shown in Stage~3 of Fig.~\ref{fig:framework}, PSD is evaluated along each sampled response prefix, placing supervision on the distribution the policy actually visits. Let $P^S_{i,g,t}$ and $P^T_{i,g,t}$ denote compact student and teacher distributions on the teacher's top 16 tokens plus one residual tail bucket. With $\beta=0.5$ and $M_{i,g,t}=\beta P^S_{i,g,t}+(1-\beta)P^T_{i,g,t}$, define
\begin{equation}
\begin{aligned}
J_{i,g,t}&=\beta D_{\mathrm{KL}}(P^S_{i,g,t}\|M_{i,g,t})\\
&\quad +(1-\beta)D_{\mathrm{KL}}(P^T_{i,g,t}\|M_{i,g,t}),\\
d_{i,g,t}&=\min\{J_{i,g,t},0.05\}.
\end{aligned}
\label{eq:token_jsd}
\end{equation}
Let $m_{i,g,t}$ mark generated tokens inside one unambiguous \texttt{<answer>} span, and let $N_{i,g}=\sum_t m_{i,g,t}$ and $N_i=\sum_gN_{i,g}$. Responses with $N_{i,g}=0$ contribute no PSD tokens. For an actor minibatch $\mathcal B$, the implementation aggregates token divergences as
\begin{equation}
\begin{aligned}
\ell_{i,g}&=N_{i,g}^{-1}\sum_t m_{i,g,t}d_{i,g,t},&&N_{i,g}>0,\\
\ell_i&=N_i^{-1}\sum_{g:N_{i,g}>0}N_{i,g}\ell_{i,g},&&N_i>0,\\
D_{\mathcal B}&=\sum_{i\in\mathcal B}z_i^{\mathrm{train}}N_i,\\
\mathcal L_{\mathrm{PSD},\mathcal B}
&=\begin{cases}
D_{\mathcal B}^{-1}\!\displaystyle\sum_{i\in\mathcal B}z_i^{\mathrm{train}}N_i\ell_i,&D_{\mathcal B}>0,\\
0,&D_{\mathcal B}=0.
\end{cases}
\end{aligned}
\label{eq:psd_aggregation}
\end{equation}
Equation~\ref{eq:psd_aggregation} is an answer-token-weighted average: rollouts and groups are not reweighted to contribute equally. The group validity threshold affects routing, but no separate per-response validity filter is applied after routing. Thus every routed response with a usable answer span participates, including a format-invalid response if that span is still unambiguous. Appendix~\ref{app:implementation} gives the exact masking behavior.

Stage~4 combines selective PSD with the GRPO objective that remains active for all groups. Let $z_{\mathcal B}^{\mathrm{train}}=\mathbf 1[D_{\mathcal B}>0]$. The complete actor loss is
\begin{equation}
\mathcal L_{\mathcal B}=\mathcal L_{\mathrm{GRPO},\mathcal B}+\lambda\,z_{\mathcal B}^{\mathrm{train}}\mathcal L_{\mathrm{PSD},\mathcal B},\qquad \lambda=0.1.
\label{eq:hybrid}
\end{equation}
Thus dense privileged guidance is selective in both groups and trajectory-answer tokens, while GRPO remains active for every group.

\subsection{Round-Wise Self-Evolution}

FIRE-VLA defines the sequence $\pi_0\rightarrow\pi_1\rightarrow\cdots\rightarrow\pi_K$. Within round $k$, a frozen privileged copy of $\pi_k$ teaches a trainable student initialized from the same parameters. Training applies Eq.~\ref{eq:hybrid} to on-policy groups sampled from the fixed RL prompt distribution. The updated student becomes $\pi_{k+1}$ and produces a new rollout distribution. Its unresolved groups then determine where privileged supervision is applied in the next round. The repeating unit is
\begin{equation}
\begin{aligned}
\operatorname{Unresolved}(\pi_k)&\longrightarrow \text{privileged supervision}\longrightarrow\pi_{k+1}\\
&\longrightarrow \operatorname{Unresolved}(\pi_{k+1}).
\end{aligned}
\end{equation}
In this way, unresolved failures of $\pi_k$ provide privileged supervision for $\pi_{k+1}$. Our experiment instantiates $K=2$ with 75 updates per round.


\section{Experiments}
\label{sec:experiments}

\subsection{Experimental Setup}

\paragraph{Training comparison.}
Both methods start from the same Qwen2.5-VL-3B SFT checkpoint. Standard GRPO runs continuously for 150 policy updates. FIRE-VLA uses two 75-update evolution rounds. A frozen 1,200-prompt set is sampled without replacement and partitioned into disjoint 600-prompt subsets for FIRE-VLA Rounds 1 and 2. Each subset is shuffled without replacement, so every FIRE-VLA prompt is used once. GRPO shuffles the full 1,200-prompt set and also uses each prompt once. Each method samples four responses per prompt, totaling 4,800 on-policy rollouts. Full-parameter BF16 training uses four RTX~3090 GPUs, FSDP, learning rate $5\times10^{-7}$, weight decay $10^{-2}$, global actor batch size 8, and a 384-token response limit. This design matches initialization, unique-prompt set, rollout count, and policy-update count, but not minibatch order. FIRE-VLA incurs additional frozen-teacher forward passes, and its second round restarts optimizer and scheduler state; it is therefore not an equal-compute comparison. Complete settings and routing incidence appear in Appendix~\ref{app:training}.

\paragraph{Evaluation protocol.}
The frozen evaluation contains 6,019 examples from 150 scene-disjoint nuScenes scenes \cite{caesar2020nuscenes}, with no scene overlap with RL or SFT training. Every checkpoint uses the same ordered samples. We evaluate one low-temperature response ($n=1$, temperature 0.2) and four stochastic responses ($G=4$, temperature 0.8). The former remains a sampled output and is termed \emph{single-sample low-temperature}, not deterministic. For $G=4$, reward and L2 are computed for every candidate, averaged within sample, and then averaged across samples.

All GRPO--FIRE comparisons are paired by sample. Confidence intervals use 10,000 bootstrap replicates with scene as the resampling cluster. The evaluation-time persistent detector is the frozen SFT-derived criterion defined in Sec.~\ref{sec:method}, not the batch-relative training route. Reproducibility and leakage checks are reported in Appendix~\ref{app:repro}.

\subsection{Main Planning Results}

\begin{table}[t]
\centering
\caption{Single-sample low-temperature planning results on the frozen 6,019-sample evaluation set. Lower is better for all L2 metrics.}
\label{tab:nominal}
\small
\setlength{\tabcolsep}{3.5pt}
\resizebox{\columnwidth}{!}{%
\begin{tabular}{lccccc}
\toprule
Method & Reward $\uparrow$ & L2@1s $\downarrow$ & L2@2s $\downarrow$ & L2@3s $\downarrow$ & Avg. L2 $\downarrow$ \\
\midrule
Standard GRPO & 0.6788 & 0.2135 & 0.6396 & 1.5302 & 0.6421 \\
FIRE-VLA & 0.6785 & 0.1997 & 0.6106 & 1.3898 & 0.6023 \\
\bottomrule
\end{tabular}}
\end{table}


Table~\ref{tab:nominal} evaluates the single-sample low-temperature regime. FIRE-VLA has lower numerical error at all three horizons and reduces Avg. L2 from 0.642 to 0.602~m, while reward is nearly unchanged. The scene-clustered paired interval for Avg. L2 is $[-0.1104,0.0087]$~m and crosses zero. We therefore interpret the two policies as having comparable nominal planning rather than a statistically conclusive FIRE-VLA advantage.

\begin{table}[t]
\centering
\caption{G=4 stochastic evaluation. Candidate metrics are averaged within sample before dataset aggregation.}
\label{tab:stochastic}
\small
\resizebox{\columnwidth}{!}{%
\begin{tabular}{lcccc}
\toprule
Method & Reward $\uparrow$ & Avg. L2 $\downarrow$ & Persistent $\downarrow$ & Any $>10$m $\downarrow$ \\
\midrule
Standard GRPO & 0.6370 & 1.8478 & 784/6019 (13.03\%) & 1.35\% \\
FIRE-VLA & 0.6156 & 1.5001 & 674/6019 (11.20\%) & 0.83\% \\
\bottomrule
\end{tabular}}
\end{table}


Repeated sampling reveals a clearer difference between the learned policy distributions (Table~\ref{tab:stochastic}). FIRE-VLA lowers $G{=}4$ Avg. L2 from 1.848 to 1.500~m, an 18.8\% reduction. The paired difference is $-0.3476$~m with scene-clustered 95\% CI $[-0.7711,-0.0319]$. This reduction is concentrated in rare severe rollouts rather than distributed uniformly across ordinary trajectories. Standard GRPO retains the higher scalar reward, 0.6370 versus 0.6156; Appendix~\ref{app:audit} examines this reward--geometry discrepancy.

\subsection{Learning from Unresolved Failures}

The separate SFT calibration split fixes a cross-model persistent-failure detector before comparison. This evaluation detector is distinct from the training-time unresolved-failure route. Applied to the complete evaluation, it marks 784/6,019 samples as persistent for GRPO (13.03\%) and 674/6,019 for FIRE-VLA (11.20\%). The $-1.83$ percentage-point paired difference has scene-clustered 95\% CI $[-2.55,-1.11]$. FIRE-VLA therefore leaves fewer samples under the frozen evaluation criterion.

The most pronounced difference appears in severe stochastic failures, although FIRE-VLA does not explicitly optimize tail risk. Samples containing at least one candidate above 10~m decrease from 81/6,019 (1.35\%) to 50/6,019 (0.83\%). The paired difference is $-0.52$ percentage points, with scene-clustered 95\% CI $[-0.83,-0.22]$. Appendix~\ref{app:additional} reports the complete distributional statistics, and Appendix~\ref{app:tail} gives the winsorized analysis.

We define recovery as an SFT-reference-persistent sample becoming negative under the same frozen evaluation detector. On the fixed set of 484 SFT reference-persistent failures, GRPO recovers 113 (23.35\%) and FIRE-VLA recovers 122 (25.21\%). This recovery-rate difference is not statistically conclusive. FIRE-VLA reduces current evaluation-persistent prevalence, but the available evidence does not establish uniformly better recovery of every original SFT failure.

\subsection{Qualitative Analysis}

\begin{figure*}[!t]
    \centering
    \includegraphics[width=.96\textwidth]{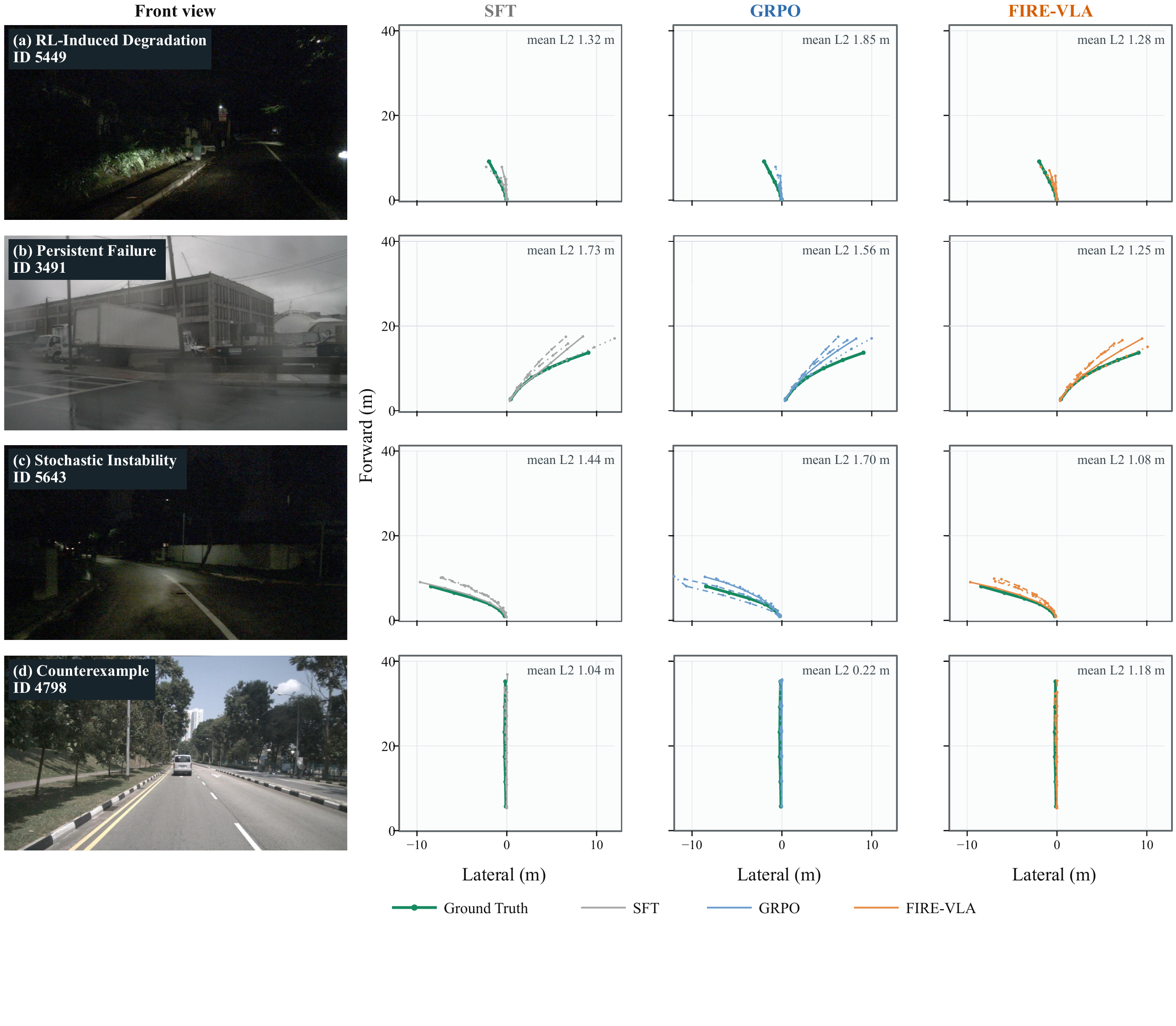}
    \caption{\textbf{Qualitative evidence on four distinct scenes.} Rows show (a) RL-induced degradation (sample 5449), (b) an evaluation-persistent failure (sample 3491), (c) non-catastrophic stochastic instability (sample 5643), and (d) a counterexample (sample 4798). In row (b), GRPO satisfies the frozen evaluation-persistent criterion, whereas FIRE-VLA does not. Each row shares the front view, ground-truth trajectory, coordinate frame, and plotting extent across methods. All four stochastic candidates are shown; no candidate selection is used.}
    \label{fig:qualitative}
\end{figure*}

Figure~\ref{fig:qualitative} visualizes four complementary regimes relevant to the proposed mechanism. In row (a), GRPO increases the group mean L2 relative to SFT, whereas FIRE-VLA restores it to the SFT error range. Row (b) shows a roadway-direction-audited example under the frozen evaluation detector: GRPO remains positive and FIRE-VLA becomes negative. Row (c) illustrates reduced non-catastrophic stochastic dispersion and lower mean L2 under FIRE-VLA. Row (d) deliberately preserves a counterexample in which GRPO is better. All plots show the complete $G{=}4$ rollout groups rather than selected candidates. Appendix~\ref{app:qualitative} describes the fixed selection protocol.

Parsing, coordinate, and repeatability checks are reported in Appendix~\ref{app:evaluation}.


\section{Discussion}

\paragraph{Role of failure-informed routing.}
GRPO and privileged distillation need not compete uniformly. Groups with varied outcomes already provide useful relative advantages, so standard GRPO remains the appropriate update. In an unresolved low-reward, low-diversity group, GRPO can still rank failures, but their relative ordering need not identify a behavior that escapes the failed region. FIRE-VLA uses that joint statistic as a switch for additional supervision while preserving the RL objective for every sample. The gate therefore responds to limited corrective information rather than assuming that low diversity makes the GRPO gradient vanish.

\paragraph{Role of the same-policy teacher.}
Student and teacher begin each round with identical weights and the same parameter scale. The teacher is stronger only because its context includes the hidden future trajectory. Conditioning both branches on the student's generated prefix places supervision on response states visited by the current policy. No larger external model or externally generated correction is required, and privileged information is removed with the teacher branch at deployment.

\paragraph{Role of round-wise self-evolution.}
A fixed teacher encodes a fixed view of failure. FIRE-VLA instead promotes and resamples the updated policy after each round. The teacher policy and the on-policy distribution of unresolved failures therefore change together. Failures of $\pi_t$ determine the privileged supervision received by $\pi_{t+1}$. Our experiment evaluates two rounds and does not establish monotonic improvement beyond them.

\paragraph{Limitations.}
The evidence supports fewer evaluation-persistent and severe stochastic failures, while nominal planning remains comparable. One training run per method leaves variation across training seeds unmeasured by the evaluation intervals. FIRE-VLA also adds teacher-forward compute and restarts optimizer and scheduler state between rounds, so the comparison is not equal-compute. The end-to-end comparison does not isolate failure routing, privileged distillation, and teacher promotion as separate components. Moreover, the two rounds use disjoint subsets of the RL prompt pool rather than tracking identical prompts longitudinally. Finally, the front-camera evaluation is open-loop and does not measure closed-loop safety, traffic compliance, comfort, or compounding state shift.


\section{Conclusion}

FIRE-VLA addresses rollout groups in which relative reward ranks poor trajectories but provides limited guidance for leaving the failed region. A frozen, same-scale teacher uses hidden future information to supervise the student's on-policy answer tokens, and the updated policy defines the teacher and failure distribution for the next round. In the evaluated two-round setting, FIRE-VLA preserves comparable single-sample planning while reducing evaluation-persistent failures and $G{=}4$ mean trajectory error. The main benefit lies in suppressing rare severe rollouts, not in uniformly improving ordinary trajectories. These conclusions are limited to a single training seed and open-loop evaluation, and the comparison does not imply equal compute or monotonic improvement over further rounds.


\bibliographystyle{unsrt}
\bibliography{bibliography}

\appendix

\onecolumn
\section{Additional Experimental Results}
\label{app:additional}

\begin{table}[H]
\centering
\caption{Complete SFT, GRPO, and FIRE-VLA results.}
\label{tab:all_models}
\small
\begin{tabular}{llccccc}
\toprule
Mode & Method & Reward & Avg. L2 & L2@1s & L2@2s & L2@3s \\
\midrule
single-sample & Common SFT & 0.6656 & 0.6240 & 0.2050 & 0.6402 & 1.4332 \\
single-sample & Standard GRPO & 0.6788 & 0.6421 & 0.2135 & 0.6396 & 1.5302 \\
single-sample & FIRE-VLA & 0.6785 & 0.6023 & 0.1997 & 0.6106 & 1.3898 \\
G=4 & Common SFT & 0.5565 & 1.0038 & 0.3649 & 1.0048 & 2.3445 \\
G=4 & Standard GRPO & 0.6370 & 1.8478 & 0.8692 & 1.9219 & 3.9002 \\
G=4 & FIRE-VLA & 0.6156 & 1.5001 & 0.6953 & 1.5912 & 3.0597 \\
\bottomrule
\end{tabular}
\end{table}


\begin{table}[H]
\centering
\caption{Scene-clustered paired bootstrap comparisons (10,000 replicates). Deltas are FIRE-VLA minus GRPO.}
\label{tab:clustered}
\small
\begin{tabular}{lrrl}
\toprule
Metric & $\Delta$ & 95\% CI & Interpretation \\
\midrule
Single-sample Avg. L2 & -0.0397 & [-0.1104, 0.0087] & Inconclusive \\
G=4 Avg. L2 & -0.3476 & [-0.7711, -0.0319] & FIRE-VLA \\
G=4 reward & -0.0213 & [-0.0250, -0.0178] & GRPO \\
Candidate P90 & 0.0956 & [0.0721, 0.1191] & GRPO \\
Candidate P95 & 0.1224 & [0.0953, 0.1650] & GRPO \\
Candidate P99 & -0.0456 & [-0.1360, 0.0789] & Inconclusive \\
Candidate CVaR95 & -7.8035 & [-16.2990, -1.5186] & FIRE-VLA \\
Candidate CVaR99 & -39.3676 & [-81.3940, -7.8663] & FIRE-VLA \\
Worst-of-4 mean & -1.4420 & [-3.1306, -0.1835] & FIRE-VLA \\
Within-sample std & -0.6234 & [-1.3526, -0.0785] & FIRE-VLA \\
Any $>10$m rate & -0.52 pp & [-0.83, -0.22] pp & FIRE-VLA \\
Current persistent rate & -1.83 pp & [-2.55, -1.11] pp & FIRE-VLA \\
Reference recovery rate & +1.86 pp & [-1.13, 4.90] pp & Inconclusive \\
\bottomrule
\end{tabular}
\end{table}


\begin{table}[H]
\centering
\caption{Severe stochastic error statistics for G=4 rollouts.}
\label{tab:robustness}
\small
\setlength{\tabcolsep}{3.2pt}
\resizebox{.72\textwidth}{!}{%
\begin{tabular}{lccccc}
\toprule
Method & CVaR95 & CVaR99 & Worst-of-4 & Intra-std & Any $>10$m \\
\midrule
Standard GRPO & 25.52 & 118.58 & 5.57 & 2.19 & 1.35\% \\
FIRE-VLA & 17.72 & 79.21 & 4.13 & 1.56 & 0.83\% \\
\bottomrule
\end{tabular}}
\end{table}


Table~\ref{tab:all_models} adds the common SFT reference to the main GRPO--FIRE comparison. SFT defines the shared initialization and frozen reference-persistent set. Table~\ref{tab:clustered} reports every principal scene-clustered paired interval, and Table~\ref{tab:robustness} summarizes severe stochastic errors. Across 24,076 candidates per policy, GRPO/FIRE medians are 0.550/0.598~m, P90 values are 1.498/1.594~m, P95 values are 1.890/2.013~m, and P99 values are 2.926/2.881~m. The respective CVaR95 values are 25.52/17.72~m and CVaR99 values are 118.58/79.21~m. Thus GRPO is better through much of the ordinary region, while FIRE-VLA reduces the severe tail.

Relative to the common SFT initialization, both RL variants exhibit higher stochastic L2 error. FIRE-VLA mitigates a substantial portion of the degradation observed under standard GRPO, but it does not restore SFT-level stochastic robustness.

\begin{multicols}{2}

\section{Implementation Details}
\label{app:implementation}
\label{app:training}

\paragraph{Optimization.}
Both RL runs use full-parameter BF16 FSDP on four RTX~3090 GPUs. Global batch size is 8 (one micro-example per device); prompt/response limits are 3,072/384 tokens, and images contain at most 196,608 pixels. Optimization uses learning rate $5\times10^{-7}$, weight decay $10^{-2}$, warm-up ratio 0.05, and gradient-norm cap 1.0. The vision tower is trainable; LoRA and the GRPO KL penalty are disabled, and the trajectory reward is the sole scalar reward.

\paragraph{Unresolved-failure routing.}
The online gate selects the lower 30th percentiles of group reward mean and population standard deviation, subject to candidate validity of at least 0.5. It routes 56 prompt groups (224 responses) in Round~1 and 58 groups (232 responses) in Round~2. Standard GRPO remains active for all groups, while routed groups additionally receive PSD.

\paragraph{Privileged self-distillation.}
Teacher and student are copies of the same round-initial policy. The frozen teacher receives ground-truth future waypoints; the student never receives this field. Both branches condition on the same student-generated prefix. The implementation retains the teacher's top 16 tokens plus one residual probability bucket, computes Jensen--Shannon divergence with mixture weight 0.5, and caps each token divergence at 0.05 before aggregation. All responses in a routed group are eligible for PSD. Individual format validity is not a second mask: a response contributes when it contains one unambiguous answer span, including the safe truncated-response case with one opening tag and no closing tag. Missing or multiple opening tags, multiple closing tags, or an empty span produce a zero mask. The distributed actor then takes one answer-token-weighted mean over all active responses in the minibatch and applies coefficient $\lambda=0.1$. This is the aggregation formalized in Eq.~\ref{eq:psd_aggregation}.

\paragraph{Round schedule.}
The formal 1,200-prompt set was sampled without replacement with seed 42 and split by position into two disjoint 600-prompt subsets. Round~1 performs 75 updates from SFT on the first subset; its merged policy initializes the student and frozen teacher in Round~2, which performs 75 updates on the second subset. Each round uses a seeded random sampler without replacement, batch size 8, and one epoch, so every assigned prompt appears once. Standard GRPO uses the same 1,200-prompt set in a separate seeded shuffle. The two methods therefore match prompt-set membership but not update order. The empirical round logs do not track the same prompts longitudinally across FIRE-VLA rounds. The second round also restarts optimizer and scheduler state and incurs additional teacher-forward computation.

\section{Evaluation and Reproducibility}
\label{app:evaluation}
\label{app:audit}
\label{app:persistent}
\label{app:determinism}
\label{app:repro}

\paragraph{Reward, aggregation, and parsing.}
Formal training and final evaluation use the same trajectory-only reward,
\begin{equation}
r(\hat\tau,\tau^*)=\left(1+T^{-1}\sum_{t=1}^{T}\lVert \hat p_t-p_t^*\rVert_2^2\right)^{-1}.
\end{equation}
Candidate reward and L2 use the same within-sample then across-sample aggregation. We additionally audited extreme decoded trajectories and found no evidence that parsing, coordinate conversion, waypoint ordering, unit handling, or fallback behavior explains the observed large-error outputs.

\paragraph{Reward--L2 relationship.}
Formal training and final evaluation both use the bounded reciprocal transformation above. Reward therefore remains highly rank-aligned with trajectory geometry, but it approaches zero once physical error is already very large. Further increases from severe to catastrophic L2 are consequently compressed in scalar reward. Across the 6,019 sample-level $G{=}4$ aggregates per model, Spearman correlations between reward and negative Avg.~L2 are 0.9793 for GRPO and 0.9807 for FIRE-VLA, whereas Pearson correlations are 0.0781 and 0.0879. This saturation permits GRPO to have higher average scalar reward while also exhibiting a heavier extreme physical-error tail.

\paragraph{Persistent-failure detector.}
The detector was calibrated once on the 256-sample, 37-scene SFT $G{=}4$ formal validation split. This calibration split has no sample or scene overlap with the final evaluation. Among groups with validity at least 0.5, the 30th percentiles of reward mean and population reward standard deviation ($\mathrm{ddof}=0$) yielded thresholds 0.4457839238 and 0.0747977498. These values were frozen and passed unchanged to final analysis. Applying them to the 6,019-sample evaluation yields 484 SFT, 784 GRPO, and 674 FIRE-VLA persistent failures. Recovery means that a sample in the SFT-reference-persistent set becomes detector-negative under the same frozen detector; GRPO and FIRE-VLA recover 113/484 and 122/484 samples, respectively.

\paragraph{Sampling and repeatability.}
Because vLLM still samples at temperature 0.2, the single-sample setting is not strictly deterministic even with a fixed request seed. The $G{=}4$ evaluation uses temperature 0.8 and identical sampling settings for all checkpoints.

\paragraph{Split and statistical protocol.}
The final evaluation contains 6,019 samples from 150 scenes and 5,119 unique images. It has no scene overlap with RL or SFT training and no image overlap with SFT training. All GRPO--FIRE comparisons are paired by sample. Confidence intervals use 10,000 paired bootstrap replicates with scene as the resampling cluster. We release the fixed sample list, evaluation configuration, per-sample provenance, and scripts needed to reproduce all reported results.

\section{Qualitative Selection Protocol}
\label{app:qualitative}

Figure~\ref{fig:qualitative} uses four categories fixed before rendering: RL-induced degradation, persistent failure, non-catastrophic stochastic instability, and a counterexample. The same 6,019 paired samples are first filtered by category-specific criteria and normal-scale error caps. We then check agreement between the visible roadway and archived ground-truth turn before ranking outcome relevance, geometry, separation, and image quality. Distinct scenes are enforced. Every panel shares the image, crop, coordinates, ground truth, and BEV extent, and shows all four candidates for every method. No best- or worst-candidate selection is used. In the persistent row, GRPO satisfies the frozen detector and FIRE-VLA does not.

\section{Additional Distributional Analysis}
\label{app:analysis}
\label{app:tail}
\label{app:claims}

GRPO remains better at the candidate median, P90, and P95; FIRE-VLA's advantage concentrates in rare severe stochastic errors. FIRE-minus-GRPO winsorized mean differences are $+0.0464$, $+0.0440$, $+0.0372$, and $+0.0233$~m at caps of 3, 5, 10, and 20~m, versus $-0.3476$~m uncapped. Thus FIRE-VLA's lower $G{=}4$ mean L2 is associated primarily with suppressing rare severe rollouts, not uniformly improving trajectories.

\end{multicols}


\end{document}